\documentclass[pdflatex,sn-nature]{sn-jnl}

\usepackage{graphicx}
\usepackage{amsmath,amssymb,amsfonts}
\usepackage{textcomp}
\usepackage{manyfoot}
\usepackage{booktabs}
\usepackage{xcolor}
\usepackage{url}

\setcitestyle{super,sort&compress}

\title[Article Title]{Bayesian uncertainty estimation improves clinical decision making in medical AI agents}

\author*[1]{\fnm{Frederik} \sur{Hauke}}

\author[1]{\fnm{Patrick} \sur{Wienholt}}

\author[1]{\fnm{Christiane} \sur{Kuhl}}

\author[2,3]{\fnm{Dyke} \sur{Ferber}}

\author[3,4,2]{\fnm{Jakob Nikolas} \sur{Kather}}

\author[1]{\fnm{Sven} \sur{Nebelung}}

\author[1]{\fnm{Daniel} \sur{Truhn}}

\affil*[1]{\orgdiv{Department of Diagnostic and Interventional Radiology}, \orgname{University Hospital RWTH Aachen}, \orgaddress{\city{Aachen}, \country{Germany}}}

\affil[2]{\orgdiv{Department of Medical Oncology, National Center for Tumor Diseases (NCT)}, \orgname{Heidelberg University Hospital}, \orgaddress{\city{Heidelberg}, \country{Germany}}}

\affil[3]{\orgdiv{Else Kr\"oner Fresenius Center for Digital Health}, \orgname{TU Dresden}, \orgaddress{\city{Dresden}, \country{Germany}}}

\affil[4]{\orgdiv{Department of Medicine I, Faculty of Medicine and University Hospital Carl Gustav Carus}, \orgname{TUD Dresden University of Technology}, \orgaddress{\city{Dresden}, \country{Germany}}}

\abstract{Machine learning models for medical image analysis typically lack a reliable measure of confidence, limiting their use in ambiguous or atypical cases. Here we show that Monte Carlo dropout, applied to a multi-task chest-radiograph classifier (eight thoracic findings, 137{,}593 training images), provides an epistemic uncertainty signal that tracks generalisation across training-set scales and flags confident yet error-prone predictions. Adding this signal to the point prediction raised error-detection AUROC from 0.74 to 0.77 ($\Delta$AUROC +0.023, 95\,\% CI [+0.014, +0.033]). In a controlled $2\times2$ factorial experiment, a clinical-decision-support agent exploited this uncertainty only when it was delivered as a binary error-risk flag rather than as raw scores, cutting confident misdiagnoses on unreliable findings from 8.5\,\% to 2.7\,\%. Epistemic uncertainty estimation thus carries decision-relevant information beyond point predictions, but its value for downstream agents depends on how it is communicated.}


\begin{document}

\maketitle

% Introduction (unheaded, per Nature Biomedical Engineering style)

Imaging findings in clinical radiology are frequently subtle or ambiguous, particularly in early disease stages and in patients with multiple comorbidities \cite{s13244_023_01521_7}. Diagnostic uncertainty is especially common in chest radiography: radiologists routinely signal doubt through hedging language such as ``probably'', ``possible'', or ``may represent'' \cite{defined2025_clinical_uncertainty_radiology_reports}, and this hedging directly influences decisions about additional imaging and follow-up.

As machine learning models assume a growing role in medical image interpretation, an analogous capacity to represent uncertainty becomes essential. Many current models, including large language models (LLMs), produce incorrect predictions with high confidence and do not provide explicit uncertainty estimates \cite{wen2024_llm_overconfidence,cash2025_llm_confidence}. Bayesian deep-learning techniques address this shortcoming by attaching an explicit uncertainty estimate to each prediction. Their value has been demonstrated across a range of clinical tasks, including chest radiography, tuberculosis segmentation, stroke analysis and COVID-19 detection \cite{bargagna2023_uncertain_labels_cxr,rajaraman2022_tb_uncertainty,herzog2020_stroke_uncertainty,calderon2021_covid_uncertainty,bayesian_cnn_medical_scarcity}, and they have been used to flag out-of-distribution inputs and rare, underrepresented classes \cite{yang2019_bayesian_retinopathy,rezaei2023_class_imbalance}.

Two principal forms of uncertainty are commonly distinguished. Aleatoric uncertainty arises from inherent noise in the data, whereas epistemic uncertainty reflects the model's lack of knowledge and can, in principle, be reduced with more training data \cite{wimmer2023_aleatoric_epistemic}. A standard classifier trained with cross-entropy loss produces class probabilities that, in theory, approximate aleatoric uncertainty \cite{gustafsson2020_scalable_bayesian,kurz2022_uncertainty_systematic_review,gawlikowski2023_uncertainty_survey}. These probabilities cannot, however, capture epistemic uncertainty, which means they give no indication of whether a prediction falls outside the model's training experience \cite{gawlikowski2023_uncertainty_survey,gustafsson2020_scalable_bayesian}. In practice, the problem is compounded by limited and imbalanced training data, a situation common in medical imaging: the model overfits, its output probabilities become poorly calibrated, and even post-hoc corrections such as temperature scaling provide only partial relief \cite{gal2016_dropout_bayesian,gawlikowski2023_uncertainty_survey}. Separate methods are therefore needed to estimate epistemic uncertainty from the model itself.

Among such methods \cite{gawlikowski2023_uncertainty_survey}, deep ensembles demand training several independent networks \cite{lakshminarayanan2017_deep_ensembles}, whereas Monte Carlo (MC) dropout obtains uncertainty from repeated stochastic forward passes at inference time alone, offering a favourable trade-off between cost and effectiveness \cite{gustafsson2020_scalable_bayesian}. MC dropout has also been shown to yield reliable uncertainty estimates for transformer architectures \cite{shelmanov2021_transformer_uncertainty}.

Deep learning systems for chest X-ray interpretation are already in clinical use, with U.S.\ Food and Drug Administration (FDA)-cleared models achieving areas under the curve (AUCs) above 0.97 for comprehensive abnormality detection across cardiac, pulmonary and pleural findings, and improving physician accuracy \cite{anderson2024_dl_improves_cxr}. Yet an estimated 3--5\% of radiological interpretations contain errors or discrepancies \cite{brady2017_error_discrepancy_radiology}, and deploying models that cannot signal when their own predictions are unreliable risks propagating confident but incorrect outputs into clinical workflows. A systematic review concluded that the benefits of uncertainty estimation in AI-clinician collaboration remain largely uninvestigated \cite{kurz2022_uncertainty_systematic_review}.

\begin{figure}[ht]
    \centering
    \includegraphics[width=\textwidth]{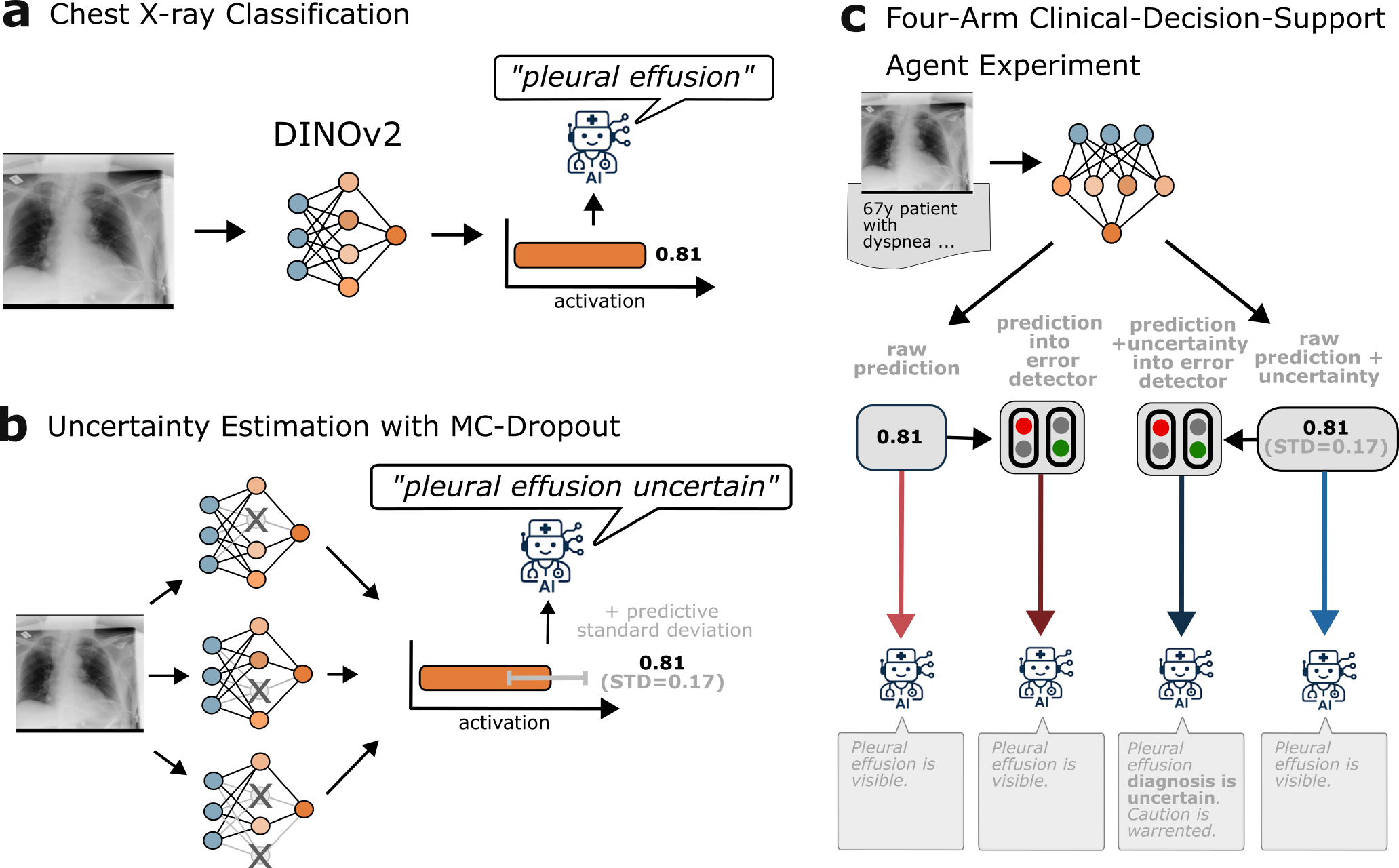}
    \caption{\textbf{Study overview.} \textbf{a}, A DINOv2 vision transformer classifies chest radiographs into eight thoracic findings, producing a single sigmoid activation per class. \textbf{b}, MC-dropout uncertainty estimation: multiple stochastic forward passes with active dropout yield the same point prediction plus a predictive standard deviation, flagging unstable outputs. \textbf{c}, Agent experiment: each held-out test case is routed to a clinical-decision-support agent under a $2\times2$ design that crosses the error-risk metric (prediction only vs prediction $+$ uncertainty) with how it is presented (raw model outputs vs a binary error-risk flag); the agent's commit-versus-escalate decisions are compared against the radiologist ground-truth label.}
    \label{fig:overview}
\end{figure}

Recent work suggests that clinical AI will increasingly rely on coordinating agents that invoke specialised models and integrate their outputs with clinical context \cite{s41586_023_05881_4,Towards_Generalist_Biomedical_AI,ferber2025_ai_agent_oncology,survey_llm_agents_medicine_2025}. Here, we equip a multi-task chest-radiograph classifier with MC-dropout uncertainty estimation. We evaluate this system in three steps: a data-scaling experiment in which models are trained on progressively smaller fractions of the data, a per-class reliability analysis, and a controlled factorial agent experiment. We find that predictive standard deviation tracks how well the model generalises, rising and falling with validation loss, and that it flags individual predictions that are confident yet unreliable. Disclosing this signal measurably improves detection of the model's own errors; a clinical-decision-support agent, however, realises this value only when the signal is delivered as a pre-digested binary flag rather than as raw numbers to interpret. An overview of the full pipeline is shown in Figure~\ref{fig:overview}.

\section{Results}

\subsection{Standard deviation co-evolves with validation loss}

\begin{figure}[ht]
    \centering
    \includegraphics[width=\textwidth]{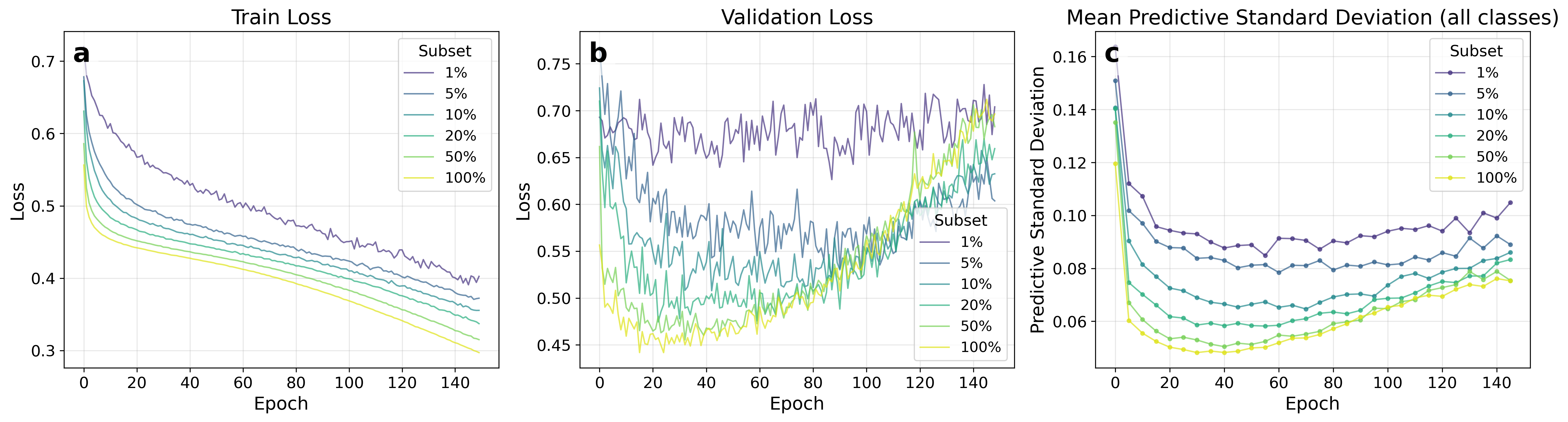}
    \caption{\textbf{Training fit, generalisation and epistemic uncertainty co-evolve across training-set size.} Curves are coloured by training-set fraction (1--100\,\%). \textbf{a}, Training loss decreases monotonically with epoch and data scale. \textbf{b}, Validation loss on the full 34{,}860-image validation set; minima are reached early and rise thereafter at every subset size. \textbf{c}, Mean MC-dropout predictive standard deviation (30 forward passes, averaged across eight findings) follows the same U-shape as validation loss, establishing standard deviation as a generalisation-dependent confidence signal.}
    \label{fig:training_curves}
\end{figure}

We first examined how training-set size influences optimisation, generalisation and epistemic uncertainty. Across the ten subset fractions, training loss decreased monotonically with epoch and with data scale, and the 100\,\% training-data run (137{,}593 radiographs) reached the lowest training loss (Figure~\ref{fig:training_curves}a). Validation loss, in contrast, was non-monotonic in epoch for every subset: it fell rapidly during the first 20--30 epochs, reached a minimum, and then rose for the remainder of training (Figure~\ref{fig:training_curves}b). The full-data run reached its minimum earliest and at the lowest value. The subsequent rise, the canonical signature of overfitting, was unambiguous and present at every data scale.

Predictive standard deviation, averaged across the eight findings on the held-out validation set, followed almost the same U-shape (Figure~\ref{fig:training_curves}c): a sharp drop over the first ${\sim}$20--30 epochs, a minimum close to the validation-loss minimum, and a steady rise thereafter. More training data lowered the standard deviation at every epoch, as expected, but it did not remove the late rise. The important point is that validation loss and predictive standard deviation rose together once the model began to overfit: MC dropout detects the loss of generalisation even when the predictions themselves remain close to 0 or 1.

\subsection{The signal generalises across all eight findings}

\begin{figure}[ht]
    \centering
    \includegraphics[width=\textwidth]{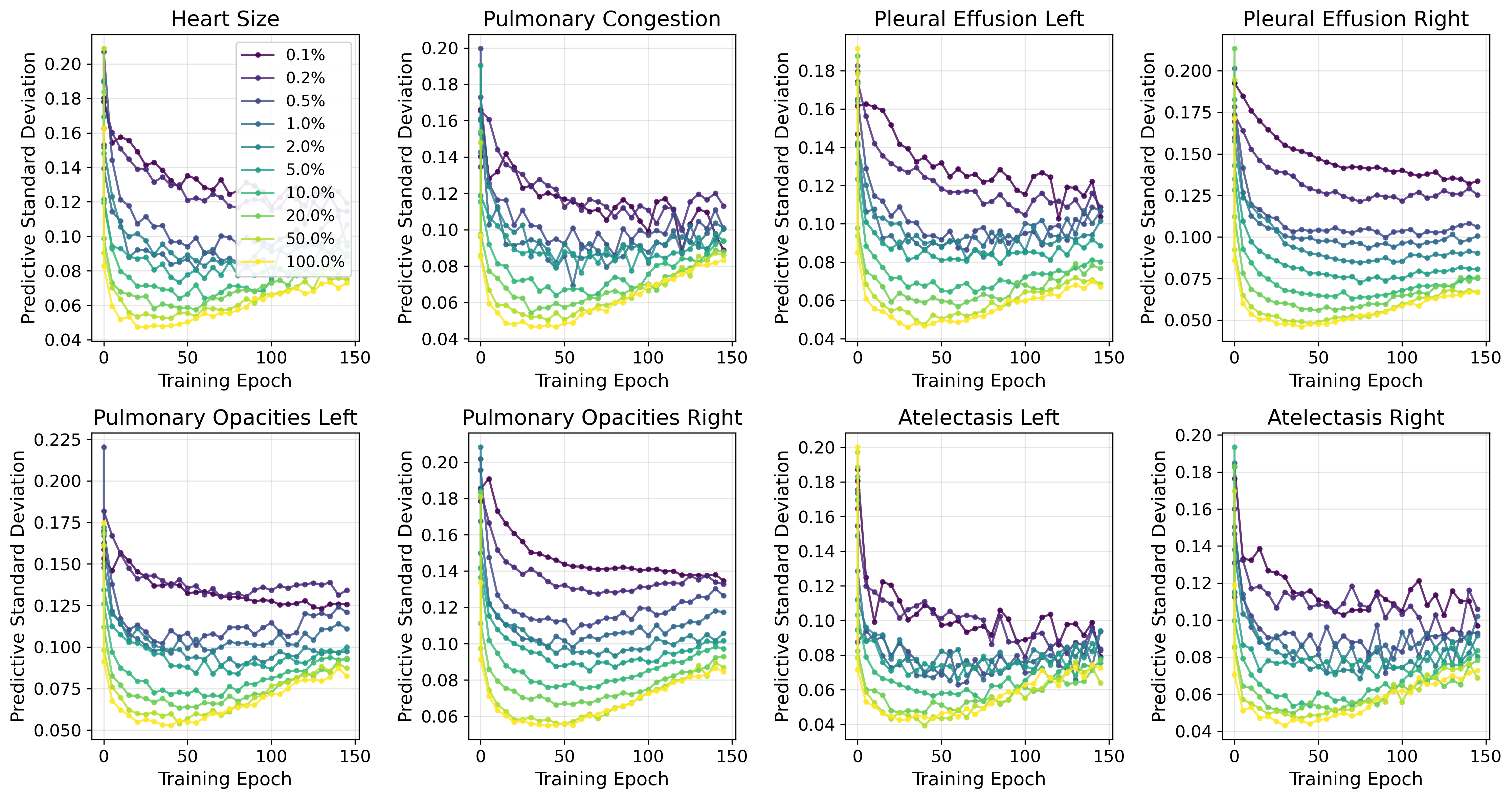}
    \caption{\textbf{The standard-deviation U-shape and data-scale ordering generalise across all eight findings.} Per-class MC-dropout predictive standard deviation on the validation set versus training epoch, coloured by training-set fraction (0.1--100\,\%). Every class reproduces the U-shape and monotonic ordering by data scale seen in Figure~\ref{fig:training_curves}c, confirming that standard deviation is a well-behaved per-class confidence signal.}
    \label{fig:perclass_variance}
\end{figure}

To verify that the late-epoch standard-deviation rise is not driven by a single anomalous class, we disaggregated predictive standard deviation by finding (Figure~\ref{fig:perclass_variance}). All eight classes exhibited the same trajectory: early collapse, a minimum near epoch 30--50, and a slow upward drift through epoch 150. The separation between subset sizes was preserved in every class. This confirms that the standard-deviation signal reflects the model's epistemic state rather than annotation noise in any particular finding.

Together, these results show that MC-dropout predictive standard deviation is a reliable confidence signal across all eight findings. It captures something the plain network output cannot: as the model starts to overfit and generalise less well, the standard deviation rises, even when the prediction itself still looks confident.

\subsection{High standard deviation flags confidently wrong predictions}

\begin{figure}[ht]
    \centering
    \includegraphics[width=\textwidth]{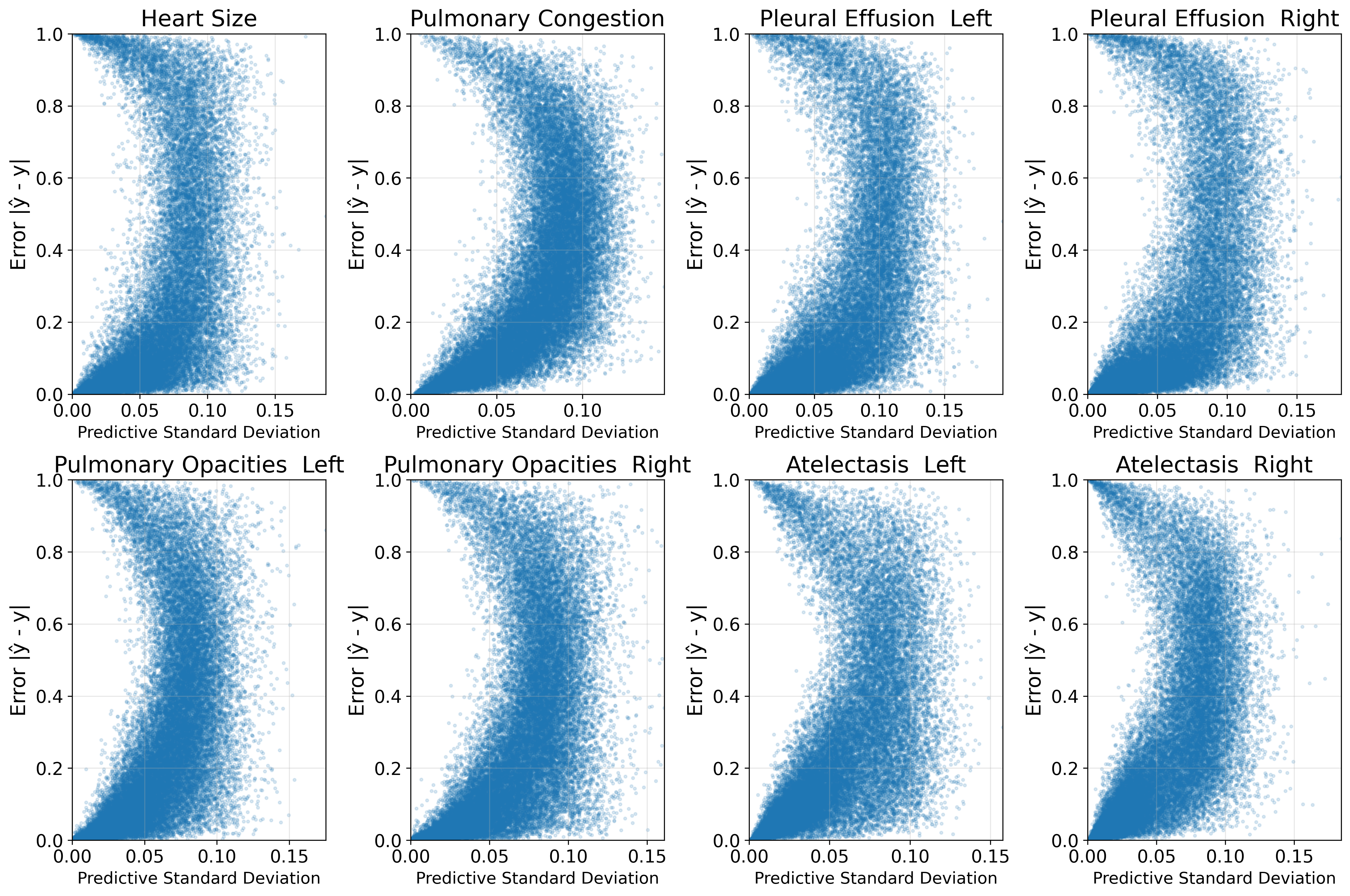}
    \caption{\textbf{Predictive standard deviation is informative about prediction errors for every finding.} Per-class scatter of absolute error $|\hat{y} - y|$ versus MC-dropout standard deviation ($n = 34{,}860$ radiographs, 30 MC samples). \textbf{a--h}, One panel per finding; predictions with high standard deviation are enriched for large errors, motivating disclosure of the uncertainty signal alongside the point prediction.}
    \label{fig:reliability_plane}
\end{figure}

We next assessed whether predictive standard deviation is informative about individual prediction errors. For each of the eight findings, we plotted the deterministic absolute error $|\hat{y} - y|$ against MC-dropout standard deviation on the full validation set ($n = 34{,}860$; Figure~\ref{fig:reliability_plane}). Most predictions clustered in the low-standard-deviation, low-error region, but every class exhibited a tail of high-error points at high standard deviation, including predictions the model reported with high confidence, where disclosing the uncertainty adds information that cannot be recovered from the point prediction. We quantify this error-detection value directly in the agent experiment below (Figure~\ref{fig:agent_experiment}).

\subsection{Uncertainty improves error detection}

\begin{figure}[ht]
    \centering
    \includegraphics[width=\textwidth]{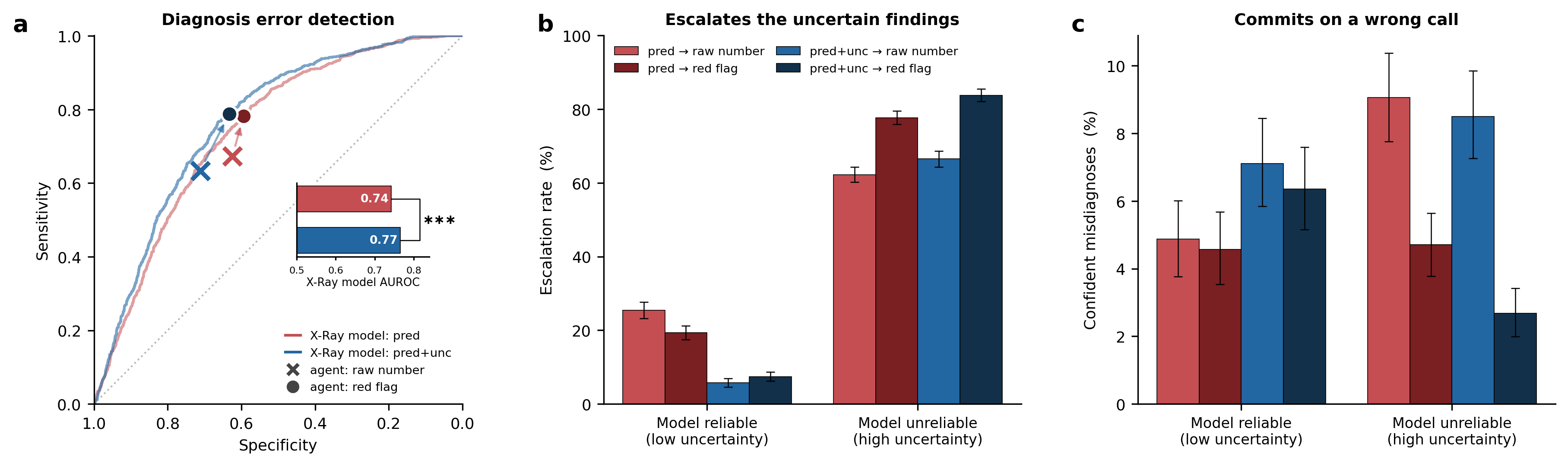}
    \caption{\textbf{The agent reaches the error-detection optimum only when the uncertainty is delivered as a binary flag.} Held-out test set; a prediction is an error when it falls on the wrong side of 0.5 relative to the radiologist ground-truth label. Two factors are crossed: the X-Ray model's error-risk metric (prediction only, red; prediction $+$ uncertainty, blue) and how it is presented to the agent (raw numbers, $\times$; binary error-risk flag, $\bullet$). \textbf{a}, Diagnosis-error-detection ROC: curves are the X-Ray model's error-ranking frontiers; markers are the agent's actual operating points (sensitivity for catching the model's errors vs specificity for passing its correct calls). Both flag agents ($\bullet$) sit on their model's frontier; both raw agents ($\times$) fall below it. Inset, X-Ray-model error-detection AUROC ($0.74$ vs.\ $0.77$; $p < 0.001$). \textbf{b}, Escalation rate, split by whether the model is reliable (below-median uncertainty) or unreliable (above-median); the flag concentrates escalation on unreliable findings. \textbf{c}, Confident misdiagnoses (the agent commits to a call that disagrees with the ground-truth label), by the same stratification; the flag reduces them on unreliable findings (prediction $+$ uncertainty $8.5\,\% \to 2.7\,\%$, $p < 0.001$) with little change where the model is reliable. Bars, mean; whiskers, case-clustered bootstrap 95\,\% CI.}
    \label{fig:agent_experiment}
\end{figure}

We first quantified, on the held-out test set, whether the uncertainty improves detection of the model's own errors beyond the point prediction alone. As described in Methods, every ordinal grade was binarised at threshold $>0$ across all splits (None negative; $(+)$ through $+++$ positive), and a prediction counted as an error when it fell on the wrong side of 0.5 relative to this binary ground truth. Averaging across all eight findings, ranking predictions by confidence alone (the distance of the sigmoid output from 0.5) detected the model's errors with an AUROC of 0.74; adding the MC-dropout uncertainty raised this to 0.77 ($\Delta$AUROC $+0.023$, 95\,\% CI [$+0.014$, $+0.033$], $p < 0.001$; case-clustered bootstrap; Figure~\ref{fig:agent_experiment}a, inset). The uncertainty signal thus carries error-relevant information that the point prediction does not.

Whether the agent could act on this information, however, depended on how it was presented. We routed each case to the agent under a $2\times2$ design that crossed the error-risk metric (prediction only vs prediction $+$ uncertainty) with its representation: either the raw per-class model outputs, which the agent must interpret itself, or a binary error-risk flag pre-computed by the model (a validation-calibrated error detector thresholded at its Youden-optimal point). Given the raw prediction-plus-uncertainty numbers, the agent's error-catching sensitivity was 0.63, leaving its operating point well below the achievable prediction-plus-uncertainty frontier and no better than an optimal use of the prediction alone (Figure~\ref{fig:agent_experiment}a). Given the same information as a binary flag, the agent reached the frontier (sensitivity 0.79), a gain of $+0.16$ (95\,\% CI $[+0.13,\, +0.18]$) over the raw representation at the prediction-plus-uncertainty level, and $+0.11$ ($[+0.08,\, +0.14]$) at the prediction-only level (both $p < 0.001$). The agent could not optimally interpret the raw model outputs, but followed the pre-digested decision to the model's own error-detection optimum.

This sharpened the agent's caution where it mattered. Splitting each finding by whether the model was reliable (below-median uncertainty) or unreliable (above-median), the flag concentrated escalation on the unreliable findings (prediction $+$ uncertainty: $66\,\% \to 84\,\%$ on unreliable, with little change on reliable; Figure~\ref{fig:agent_experiment}b). On those unreliable findings, the fraction of confident misdiagnoses (cases where the agent committed to a call that disagreed with the radiologist ground-truth label) fell from $8.5\,\%$ with the raw numbers to $2.7\,\%$ with the flag ($-5.8$ percentage points, $p < 0.001$), with little change on reliable findings (Figure~\ref{fig:agent_experiment}c).

\subsection{Agents struggle to judge raw uncertainty}

The agent's shortfall was therefore one of interpretation, not of information: the identical error-relevant signal was present in both representations, yet only the formalized flag let the agent reach the model's error-detection optimum, whereas the raw numbers left it well short. This points to a design principle for agentic clinical AI: the numerical reasoning belongs in the specialist model, which should hand the agent a calibrated, pre-digested decision rather than raw scores to interpret.

\section{Discussion}

We have shown that equipping a medical image classifier with MC-dropout uncertainty estimation and disclosing that signal to a clinical-decision-support agent yields measurably better decisions, provided the signal is delivered in a form the agent can act on. Our approach couples a vision-transformer classifier that attaches an epistemic uncertainty estimate to every prediction with a reliability-plane analysis showing that predictions with high standard deviation are enriched for errors; a controlled factorial experiment on a held-out test set then demonstrated that this uncertainty improves detection of the model's own errors, but that the agent captures the gain only when the error risk is pre-digested into a binary flag rather than handed over as raw numbers to interpret.

The agent experiment is, to our knowledge, the first controlled demonstration that the representation of a model's uncertainty, not merely its availability, governs whether an AI agent can use it in a simulated clinical-support setting. Because the four conditions were identical except for the error-risk metric and its representation, the shift in the agent's operating point (Figure~\ref{fig:agent_experiment}) is attributable to those factors alone. Handed the raw model outputs, the agent under-used the signal and stayed below the achievable frontier; handed a binary flag, it selectively escalated the findings the model was unsure about, ordering verification steps such as a radiologist second read where they were warranted while leaving confident, reliable calls untouched, and thereby reduced confident misdiagnoses on unreliable findings without over-escalating reliable ones.

More broadly, as language-model agents are increasingly placed downstream of quantitative predictors, this argues for keeping the numerical reasoning in the specialist model and handing the agent a calibrated, formalized decision; a consideration that extends beyond uncertainty to any continuous model output.

The data-scaling and per-class analyses provide the mechanistic foundation for this result. Predictive standard deviation followed validation loss closely across training-set sizes (Figure~\ref{fig:training_curves}), rising during overfitting even as the deterministic predictions stayed pinned near 0 or 1. This pattern held consistently across all eight findings (Figure~\ref{fig:perclass_variance}), confirming that MC dropout tracks generalisation degradation at the per-class level. The reliability-plane analysis (Figure~\ref{fig:reliability_plane}) then translated this aggregate signal into case-level risk stratification, flagging predictions whose instability under dropout is not apparent from the point prediction alone. These findings connect to prior work on uncertainty-based out-of-distribution detection \cite{yang2019_bayesian_retinopathy,rezaei2023_class_imbalance} and to the broader challenge of overconfidence in both human decision making and large language models \cite{wen2024_llm_overconfidence}.

Our study has several limitations. Our evaluation used a single imaging modality and a single institutional cohort; extension to other modalities, pathologies and external datasets is needed to establish generalisability. MC dropout, while practical, could be complemented or replaced by more expressive methods such as deep ensembles \cite{lakshminarayanan2017_deep_ensembles} or variational inference \cite{kingma2013_vae}. The agent's commit-versus-escalate decisions were read from a structured field it emitted rather than adjudicated by clinician assessors: the fully automated pipeline evaluated 2{,}000 agent consultations (500 cases $\times$ four conditions; ${\sim}$16{,}000 finding-level decisions), a sample size out of scope for clinician adjudication. A formal clinical validation therefore remains future work, but because the uncertainty advantage acts on the minority of cases where the model is unreliable, a clinical deployment operating at scale would encounter many more such cases and stands to benefit correspondingly more.
Looking ahead, these results argue for integrating uncertainty-aware models into AI-assisted medical image analysis, particularly as clinical workflows increasingly delegate interpretation to autonomous AI agents \cite{ferber2025_ai_agent_oncology}. We have shown that a model's apparent confidence and the stability of its predictions can come apart, and that measuring this stability requires nothing more than repeated forward passes at inference time. This makes uncertainty disclosure a concrete, readily deployable signal that can inform downstream decision support, though its clinical value remains to be validated with human clinicians.

\section{Methods}

\subsection{Dataset, splits and label schema}

We used the publicly available TAIX-Ray Thorax chest-radiograph cohort \cite{taixray2026_dataset}. All experiments used a patient-stratified split comprising 137{,}593 training, 34{,}860 validation and 42{,}928 test radiographs, with no patient appearing in more than one split. The data-scaling and reliability analyses were performed on the validation set; the agent experiment was performed on the held-out test set. Every radiograph carries an ordinal annotator grade for eight findings. HeartSize is graded on a four-level scale (Normal, Borderline, Enlarged, Massively enlarged), while the remaining seven findings, Pulmonary Congestion, Pleural Effusion (Left/Right), Pulmonary Opacities (Left/Right), and Atelectasis (Left/Right), each use five grades (None / (+) / + / ++ / +++). For the training objective the grades were binarised at threshold $> 0$; the raw 0-4 grades were preserved on disk and re-coupled to inference outputs post-hoc at analysis time, including the annotator-uncertain label (grade~1: Borderline = ``(+)'').

Radiographs were centre-cropped to $448 \times 448$\,px and per-image z-normalised. Training augmentations included random cropping, vertical flipping and axis transposition; no augmentations were applied at validation or test time.

\subsection{DINOv2 backbone with MC dropout}

The classifier is a DINOv2 Vision Transformer (ViT)-S/14 \cite{oquab2023_dinov2} (${\sim}$21\,M parameters), fine-tuned from the public self-supervised checkpoint on the eight binarised labels via a single linear head. The training objective was per-class binary cross-entropy; optimisation used AdamW (learning rate $1 \times 10^{-6}$, weight decay $1 \times 10^{-2}$) in mixed-precision 16-bit floating point (FP16).

For epistemic uncertainty estimation, all dropout layers inside the transformer were enabled at inference time with rate $p = 0.1$, and an additional dropout layer was inserted before the linear head. Throughout this paper we report MC-dropout estimates from 30 stochastic forward passes per image \cite{gal2016_dropout_bayesian}.
Per-class predictive standard deviation is the standard deviation of the sigmoid-space MC samples; per-class predictive entropy is the binary entropy of the MC mean. The deterministic prediction is a single forward pass with dropout disabled, decoupling what the model predicts from how stable that prediction is under stochastic perturbation.

\subsection{Data-scaling experiment}

To examine the joint behaviour of fit, generalisation and epistemic uncertainty as a function of training-set size, we trained ten independent classifiers on nested random subsets of the training split (0.1--100\,\%, spanning ${\sim}$138 to 137{,}593 radiographs). Each run lasted 150 epochs with batch size 48 and identical hyper-parameters; the validation set was always the full 34{,}860-image split.

Every five epochs, 30-sample MC-dropout inference was run on the full validation set and the mean predictive standard deviation, entropy and prediction were recorded per class (Figure~\ref{fig:training_curves}, Figure~\ref{fig:perclass_variance}).

\subsection{Reliability analysis on the validation set}

For the reliability analysis we used a single converged checkpoint (epoch~37) and ran 30-sample MC-dropout inference on the full validation set (34{,}860 radiographs). For each image-class pair we recorded the deterministic prediction, MC-dropout standard deviation, binary target and absolute error $|\hat{y} - y|$ (Figure~\ref{fig:reliability_plane}). A prediction was counted as an error when it fell on the wrong side of 0.5 relative to the binarised ground-truth label.

\subsection{Factorial agent experiment}

For the agent experiment we drew a representative sample of 500 cases uniformly at random (fixed seed) from the held-out test split, yielding $500 \times 8 = 4{,}000$ finding-level predictions (493 cases had all four conditions successfully parsed). For each case GPT-5.1 (OpenAI) generated a realistic clinician request note from the image and ground-truth grades, without reference to imaging findings.

Each case was routed to a GPT-5.1-based clinical-decision-support agent under a $2\times2$ design crossing the error-risk metric with its representation. The metric was either the prediction alone or the prediction together with MC-dropout uncertainty. The representation was either raw (the per-class deterministic prediction, and, for the prediction-plus-uncertainty metric, the MC-dropout standard deviation with a per-class orientation block (validation-set minimum, 75th percentile, maximum)) or a binary error-risk flag. The flag was produced by a logistic-regression error detector (on standardised confidence for the prediction-only metric, or on standardised confidence and uncertainty for the prediction-plus-uncertainty metric) fitted on the validation set and thresholded at its Youden-optimal operating point, marking each finding ``elevated risk'' or ``normal''. All four prompts were byte-identical except for this data block. The rubric asked the agent to decide, for each finding, to commit (accept the model's call and fold it into decisive management) or escalate (order a specific confirmatory or better-suited next step, e.g.\ radiologist second read, point-of-care ultrasound, echocardiography, repeat radiograph or CT).

For each finding in each condition, the agent emitted its decision (commit or escalate) in a structured output field. A finding was counted as a confident misdiagnosis when the agent committed to a call that disagreed with the ground-truth label. Findings were stratified as reliable or unreliable by splitting each finding at its median MC-dropout uncertainty across the sample.

The X-Ray model's error-detection frontiers (Figure~\ref{fig:agent_experiment}a) are the ROC of two scores against the ground-truth error label: prediction only ($-|\hat{y} - 0.5|$) and prediction plus uncertainty (the out-of-sample probability of a 5-fold cross-validated logistic regression on the standardised confidence and uncertainty). The agent's actual decisions define operating points (sensitivity, specificity) on the same axes.

\subsection{Statistics}

Confidence intervals and $p$ values were obtained by a case-clustered bootstrap that resamples cases (3{,}000 resamples), respecting the non-independence of the eight findings within a case. The $p$ value for the error-detection AUROC difference (prediction-plus-uncertainty minus prediction-only) is the two-sided bootstrap tail probability. The error bars on the escalation-rate and confident-misdiagnosis panels (Figure~\ref{fig:agent_experiment}b,c) are the 95\,\% confidence interval of this bootstrap. No case was excluded post hoc. P values below 0.001 are reported as $p < 0.001$.

\backmatter

\bmhead{Acknowledgements}

The authors acknowledge the support of the European High Performance Computing Joint Undertaking (JU) under grant agreement No 101250682.

\section*{Declarations}

\begin{itemize}
\item \textbf{Funding} Funded by the European Union. This work received funding from the European High Performance Computing Joint Undertaking (JU) under grant agreement No 101250682 (JAIF). DT is supported by the German Ministry of Research, Technology and Space (TRANSFORM LIVER -- 031L0312C, DECIPHER-M -- 01KD2420B), the Deutsche Forschungsgemeinschaft (DFG, 515639690), and the European Union (Horizon Europe, ODELIA -- GA 101057091, ERC Starting Grant SAGMA -- GA 101222556).
\item \textbf{Conflict of interest} DT holds shares in StratifAI and Synagen. He has received honoraria from Bayer, AstraZeneca, Philips, Roche, Pfizer, and Gilead. JNK declares ongoing consulting services for AstraZeneca, Panakeia, and Bioptimus. Furthermore, he holds shares in StratifAI, Synagen, and Spira Labs, has received an institutional research grant from GSK and AstraZeneca, as well as honoraria from AstraZeneca, Bayer, Daiichi Sankyo, Eisai, Janssen, Merck, MSD, BMS, Roche, Pfizer, and Fresenius. DF holds shares and is an employee at Synagen and has received a research grant from OpenAI. All other authors declare no conflicts of interest.
\item \textbf{Availability of data and materials} The UKA Thorax chest-radiograph cohort (taIX-Ray) is publicly available \cite{taixray2026_dataset}.
\item \textbf{Code availability} The code is openly available at \url{https://github.com/TruhnLab/BayesianCXRAgent}.
\item \textbf{Authors' contributions} Frederik Hauke: Investigation, Formal Analysis, Writing -- Original Draft. Patrick Wienholt: Writing -- Review \& Editing. Christiane Kuhl: Writing -- Review \& Editing. Dyke Ferber: Writing -- Review \& Editing. Jakob Nikolas Kather: Writing -- Review \& Editing. Sven Nebelung: Writing -- Review \& Editing. Daniel Truhn: Conceptualization, Writing -- Review \& Editing.
\end{itemize}

\end{document}